\documentclass[runningheads]{llncs}
\usepackage[T1]{fontenc}
\usepackage{graphicx,verbatim}
\usepackage[table]{xcolor}
\usepackage{multirow}
\usepackage{makecell}
\usepackage{pgf}
\usepackage[colorlinks,citecolor=blue]{hyperref}
\usepackage{amsmath, amssymb}
\usepackage[british]{babel}
\usepackage{todo}
\usepackage{enumitem}
\usepackage{marvosym}

\begin{document}
\title{Harnessing Adversarial Distillation to Customise Debiased, Disease-Specific Pathology Foundation Models for Breast Cancer}
\titlerunning{Adversarial Distillation for Debiased Breast Cancer Foundation Models}
%

\author{
Zhiwei Chen\inst{1}\thanks{Zhiwei Chen and Yang Hu contribute equally to this work.\\
Corresponding email: \email{superhy199148@hotmail.com} and
\email{yangkx@scut.edu.cn}} \and
Yang Hu\inst{2,3,4\star}\textsuperscript{(\Letter)} \and
Yuxiang Xiao\inst{1} \and
Yakun Ju\inst{2} \and
Tianyang Zhang\inst{4} \and
Yingxue Xu\inst{6} \and
Wei Li\inst{7} \and
Hao Chen\inst{6} \and
Jens Rittscher\inst{4,5} \and
Kaixiang Yang\inst{1}\textsuperscript{(\Letter)}
}


\authorrunning{Zhiwei Chen, Yang Hu and Yuxiang Xiao et al.}

\institute{
School of Computer Science and Engineering, South China University of Technology, China
\and
School of Computing and Mathematical Sciences, University of Leicester, UK
\and
Leicester Cancer Research Centre, University of Leicester, UK
\and
Department of Engineering Science, University of Oxford, UK
\and
Nuffield Department of Medicine, University of Oxford, UK
\and
Department of Computer Science and Engineering, The Hong Kong University of Science and Technology, Hong Kong SAR, China
\and
ZoyMed, China
}

\maketitle              
\setcounter{footnote}{0}  
\begin{abstract}
Pathology foundation models (PFMs) provide strong tissue representations and have become central to digital pathology. However, deployment in disease-specific settings is limited by 1) the high computational cost of billion-parameter PFMs and 2) distribution mismatch and non-biological bias inherited from pan-cancer, multi-centre pre-training, including site-specific signatures and imbalanced disease prevalence. These factors can encourage shortcut learning and under-emphasise subtle morphology required for reliable modelling of a specific cancer type. We present \textbf{SmartStu} (a \textbf{Smart} \textbf{Stu}dent), a framework to customise compact, breast-cancer-specific PFMs via distillation whilst mitigating confounding.
SmartStu distils representations from multiple teacher PFMs into a lightweight student backbone. Crucially, we introduce \textbf{adversarial distillation} that leverages a dedicated noise model trained to predict nuisance, edge-dominated cues on the distillation set. Using this noise model as a counterexample, the adversarial objective encourages the student to recognise, yet suppress, features predictive of nuisance targets. We further incorporate multi-teacher ensemble distillation and an auxiliary self-supervised objective with artefact injection.
We validate SmartStu on three external cohorts (Yale HER2, SLN-Breast, and BRACS) with multiple tiny backbones. SmartStu yields breast-cancer-specific PFMs that are over $30\times$ smaller than general PFMs whilst largely preserving, and sometimes improving, downstream performance measured by balanced accuracy (bAcc) and AUC. Code is available at \url{https://github.com/zwchen03/advDistall}.



\end{abstract}

\section{Introduction}

Self-supervised pre-training has driven the rapid emergence of pathology foundation models (PFMs), enabling strong and transferable representations for histology across a wide range of downstream tasks \cite{caron2021emerging_dino,oquab2023dinov2,zhou2021ibot,chen2024uni_uni,vorontsov2024foundation_virchow,filiot2024phikonv2largepublicfeature_phikonv2,zimmermann2024virchow2_virchowv2}. Pre-trained on large, heterogeneous collections of whole-slide images (WSIs), these models capture broad morphological priors that outperform classic histopathology feature extraction approaches limited by annotation cost and dataset scale \cite{campanella2019clinical_SLNbreast,coudray2018classification_sup_work}. Despite these advances, deploying PFMs in disease-specific clinical workflows remains challenging due to the following bottlenecks.

First, existing PFMs' parameter count and inference-time cost hinder high-throughput processing and deployment in resource-constrained clinical environments \cite{li2025survey,xiong2025survey}. Scaling PFMs towards whole-slide modelling and ultra-long context \cite{xu2025multimodal,xu2024whole_Prov-GigaPath} typically increases memory use and latency, restricting deployment to well-provisioned settings. Second, PFMs pre-trained on multi-centre datasets inevitably inherit non-biological biases from tissue preparation, staining protocols, and scanning pipelines. These site (collection source)-specific signatures can be exploited as shortcuts by downstream models, degrading out-of-distribution generalisation \cite{howard2021impact,review_stain_discrepancy,komen2024histopathological}. Finally, a task mismatch can arise from pan-cancer pre-training: PFMs optimised for broad transfer across multiple cancer types may under-emphasise subtle, cancer-specific morphology required for reliable modelling in the target disease.

An intuitive direction is to customise PFMs via knowledge distillation (KD), compressing large, general teachers into compact, task-specific students by matching representations. In histopathology, distillation can yield efficient models that preserve much of the teacher performance \cite{filiot2025distilling}. However, standard distillation largely transfers whatever the teacher encodes and offers limited control over propagating site-related bias into the student. Domain generalisation and debiasing techniques (e.g., stain normalisation) can reduce domain shift \cite{asadi2024learning,jahanifar2025domain}, but may either suppress diagnostically relevant variation or leave residual centre-identifiable structure in the representation, arising from acquisition pipelines and case-mix confounding rather than superficial stain differences.

We present \textbf{SmartStu}, a framework for customising compact, disease-specific PFMs for breast cancer by distilling from multiple teacher PFMs into a lightweight student, whilst mitigating confounding inherited from centralised pre-training. SmartStu introduces adversarial distillation: we train a dedicated noise model to predict nuisance targets and capture site-related cues on the distillation set, then use it as a counterexample during KD. We optimise an adversarial objective that encourages the student to recognise biased directions yet suppress features predictive of nuisance targets. In addition, SmartStu incorporates an auxiliary self-supervised objective with multi-type artefact injection to improve stability under acquisition variability. We summarise our contributions as:

1) We propose \textbf{SmartStu}, a disease-specific and lightweight pathology foundation model framework with an adversarial distillation mechanism that actively suppresses site-related confounding during customisation. 

2) We integrate complementary robustness components, including multi-teacher ensemble distillation and an auxiliary self-supervision with multiple artefact injection, to stabilise representation learning under realistic variability.

3) We evaluate SmartStu across 3 external breast cancer cohorts (Yale HER2, SLN-Breast, and BRACS) and various student backbones. SmartStu yields compact breast cancer PFMs that are over $30\times$ smaller than general PFMs whilst preserving, and occasionally improving, downstream performance.

\section{Methods}
\begin{figure}[t]
    \centering
    \includegraphics[width=\textwidth]{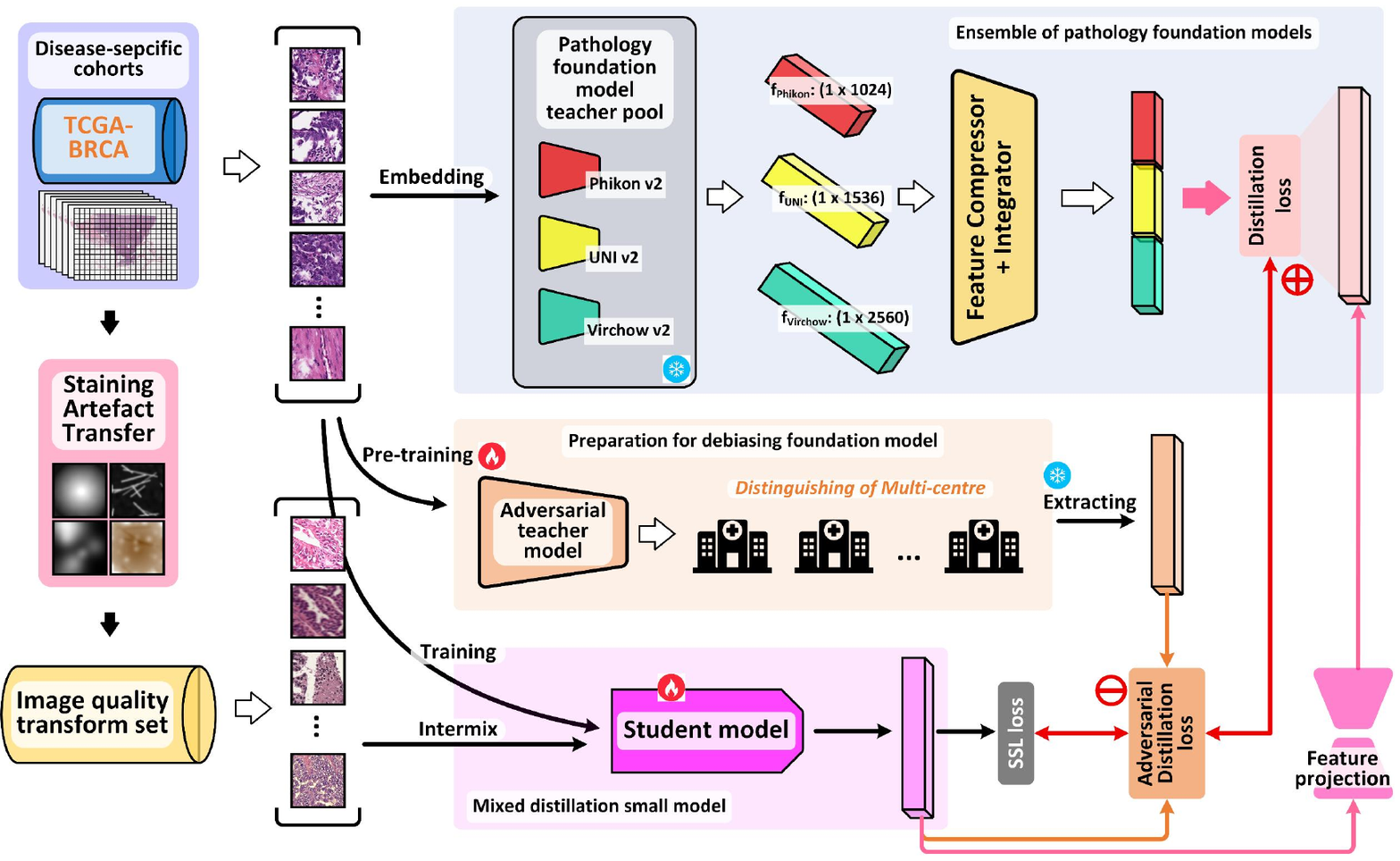}
    \caption{Overview of \textbf{SmartStu}: multi-teacher feature distillation into a compact student, augmented with \textbf{adversarial (noise) distillation} to suppress site-related confounding, and an auxiliary \textbf{self-supervised} objective with artefact injection.}
    \label{fig:method-overview}
\end{figure}

\textbf{SmartStu} is a compact tile-level disease-specific encoder and distils complementary knowledge from multiple teacher PFMs into a lightweight student while explicitly suppressing bias via adversarial distillation with a dedicated noise model (see Fig.~\ref{fig:method-overview}). Given a whole slide image (WSI) $X$, the tissue region is split into $M$ non-overlapping tiles $\mathcal{I}=\{x_i\}_{i=1}^{M}$, extracted at $20\times$ magnification with size $256\times256$ pixels. SmartStu is trained to produce domain-specific tile embeddings; weakly supervised slide-level prediction is obtained downstream by feeding the distilled embeddings into a standard MIL aggregator~\cite{ilse2018attention_ABMIL}.

\subsection{Multi-teacher feature distillation}

\label{sec:MTKD}

For each tile $x_i\in\mathcal{X}$, the student encoder produces $z_i^{s}=S(x_i)$.
To transfer diverse pathology semantics, we use a teacher pool of $K$ frozen PFMs $\{T_k\}_{k=1}^{K}$. For each $x_i\in\mathcal{X}$, teacher $T_k$ produces an embedding $f_i^{(k)}=T_k(x_i)\in\mathbb{R}^{d_k}$. Since teacher embeddings have varying dimensions, we apply a lightweight alignment head $\phi_k(\cdot)$ to map each teacher feature into a shared dimension $D$. We then form the concatenated teacher target:
\begin{equation}
z_i^{t}=\text{Concat}\Big(\phi_1(f_i^{(1)}),\phi_2(f_i^{(2)}),\dots,\phi_K(f_i^{(K)})\Big)\in\mathbb{R}^{K\cdot D}.
\end{equation}
We map student features to the teacher target space via a projection head, $h_i^{dist}=\psi_{dist}(z_i^{s})\in\mathbb{R}^{K\cdot D}$, and minimise a feature-level mean-squared error between $\{h_i^{dist}\}$ and $\{z_i^{t}\}$:
\begin{equation}
\mathcal{L}_{distill}=\frac{1}{|\mathcal{X}|}\sum_{i=1}^{|\mathcal{X}|}\left\|z_i^{t}-h_i^{dist}\right\|_2^{2}.
\end{equation}

\subsection{Adversarial distillation for de-biasing}
\label{sec:adv_distill}
SmartStu introduces adversarial distillation by learning a noise model that captures nuisance cues and then training the student to suppress them.

\noindent\textbf{Noise model pre-training (adversarial teacher).}
We pre-train an auxiliary adversarial teacher $T_{adv}$ on a multi-centre cohort with site-wise nuisance labels. For a tile $x$ with site label $y_c\in\{1,\dots,C\}$, we extract features $v=T_{adv}(x)$ and project them to a normalised embedding $z$. We optimise a hybrid objective that combines contrastive learning \cite{khosla2020supervised_SupCon} with cross-entropy ($\mathrm{CE}(\cdot)$) site classification:
\begin{align}
\mathcal{L}_{ctr} &= \sum_{i \in I} \frac{-1}{|P(i)|} \sum_{p \in P(i)} \log \frac{\exp(z_i \cdot z_p / \tau)}{\sum_{a \in A(i)} \exp(z_i \cdot z_a / \tau)},\\
\mathcal{L}_{pre} &= \mathrm{CE}(y_c,p_c) + \lambda\,\mathcal{L}_{ctr}.
\end{align}
Here $P(i)$ denotes positives from the same site, $A(i)$ contains all samples in the batch, $\tau$ is a temperature, and $p_c$ is the predicted site distribution. After pre-training, $T_{adv}$ is frozen and used as a nuisance-feature extractor.

\noindent\textbf{Adversarial distillation loss.}
For each $x_i\in\mathcal{X}$, the frozen noise teacher outputs a nuisance signature $z_i^{adv}=T_{adv}(x_i)$. The student feature $z_i^{s}$ is mapped by a feature alignment head $\psi_{adv}(\cdot)$ and then passed through a gradient reversal layer (GRL) \cite{ganin2016domain_GRL}, giving $h_i^{adv}=\mathcal{R}\!\left(\psi_{adv}(z_i^{s})\right)$. We use cosine similarity (bounded in $[-1,1]$) to measure alignment between $h_i^{adv}$ and $z_i^{adv}$:
\begin{equation}
\mathcal{L}_{adv}=1-\text{CosSim}(z_i^{adv},h_i^{adv})
=1-\frac{z_i^{adv}\cdot h_i^{adv}}{\|z_i^{adv}\|_2\|h_i^{adv}\|_2}.
\end{equation}
With GRL, minimising $\mathcal{L}_{adv}$ discourages the student from encoding site-related confounders in its representation.

\subsection{Auxiliary self-supervised adaptation \& overall optimisation}
\noindent\textbf{Training batch with artefact injection.}
We apply an artefact injection operator $\mathcal{A}(\cdot)$ during training, including 1) brown colour cast to mimic stain shift, 2) exposure/illumination jitter, 3) resolution degradation via downsampling with lossy compression, and 4) additive Gaussian noise. For each mini-batch $\mathcal{B}$, we form $\mathcal{X}=\mathcal{B}\uplus\mathcal{A}(\mathcal{B})$ and optimise all objectives on $\mathcal{X}$.

\noindent\textbf{Self-supervised adaptation (SSL adaptation).}
On $\mathcal{X}$, we add a DINO-style self-supervised term \cite{caron2021emerging_dino}. For each tile, we sample two global crops and two local crops (after $\mathcal{A}(\cdot)$), and compute their predicted distributions via the student projection head and softmax. Denoting the predictions of global crops as $P^{g},P^{g}_{aug}$ and those of local crops as $P^{l},P^{l}_{aug}$, we minimise the symmetric cross-entropy:
\begin{equation}
\mathcal{L}_{ssl}=\frac{1}{2}\Big(\mathrm{CE}(P^{g},P^{l}_{aug})+\mathrm{CE}(P^{g}_{aug},P^{l})\Big).
\end{equation}



\noindent\textbf{Overall optimisation objective.}
The final training objective is a weighted combination of multi-teacher distillation, adversarial (noise) distillation, and self-supervised adaptation:
\begin{equation}
\mathcal{L}_{all}=\mathcal{L}_{distill}+\lambda_{adv}\mathcal{L}_{adv}+\lambda_{ssl}\mathcal{L}_{ssl},
\end{equation}
where $\lambda_{adv}$ and $\lambda_{ssl}$ balance the contribution of de-biasing and SSL adaptation.

\section{Experiments and Results}

\begin{table}[t]
\caption{Performance comparison when using UNI as the only teacher. Results are reported as balanced accuracy (bAcc) and AUC (mean for 5-runs). \textbf{Bold+shaded} indicates the best result, while \textbf{bold} indicates the best record for each student backbone.}
\label{tab:table1}
\fontsize{8pt}{8.6pt}\selectfont
\centering

\setlength{\tabcolsep}{0.7mm}

\begin{tabular}{@{}l l cc cc cc@{}}
\hline
\multicolumn{2}{c}{} & \multicolumn{2}{c}{Yale HER2} & \multicolumn{2}{c}{SLN-Breast} & \multicolumn{2}{c}{BRACS} \\
\cline{3-8}
Model & Mode & bAcc & AUC & bAcc & AUC & bAcc & AUC \\
\hline
UNI-v2 (681M) &  &
\cellcolor{red!12}\textbf{0.9032} & \cellcolor{red!12}\textbf{0.9500} &
\cellcolor{red!12}\textbf{0.9323} & \cellcolor{red!12}\textbf{0.9805} &
\cellcolor{red!12}\textbf{0.5083} & \cellcolor{red!12}\textbf{0.8699} \\
\hline
\multirow{4}{*}{\makecell{ResNet18\\(11.18M)}} & Default & 0.7642 & 0.8479 & 0.7707 & 0.8722 & 0.3281 & 0.7455 \\
& Sig-SmartStu & 0.8208 & 0.9105 & 0.8647 & 0.9143 & 0.3697 & 0.8083 \\
& Adv-SmartStu & 0.8361 & 0.9121 & 0.8842 & 0.9308 & 0.3984 & 0.8163 \\
& AdvS-SmartStu & \textbf{0.8468} & \textbf{0.9226} & \textbf{0.8985} & \textbf{0.9534} & \textbf{0.4378} & \textbf{0.8180} \\
\hline
\multirow{4}{*}{\makecell{TinyViT-5M\\(5.07M)}} & Default & 0.7513 & 0.8553 & 0.8323 & 0.8917 & 0.3497 & 0.7300 \\
& Sig-SmartStu & 0.8450 & 0.9192 & 0.9038 & 0.9188 & 0.4002 & 0.8016 \\
& Adv-SmartStu & \textbf{0.8553} & 0.9247 & \textbf{0.9090} & 0.9308 & 0.4088 & 0.8039 \\
& AdvS-SmartStu & 0.8453 & \textbf{0.9268} & 0.9060 & \textbf{0.9398} & \textbf{0.4119} & \textbf{0.8198} \\
\hline
\multirow{4}{*}{\makecell{MobileNetV3-Small\\(0.93M)}} & Default & 0.7582 & 0.8474 & 0.8263 & 0.8827 & 0.3313 & 0.7414 \\
& Sig-SmartStu & 0.8103 & 0.9011 & 0.8752 & 0.9218 & 0.3618 & 0.7529 \\
& Adv-SmartStu & \textbf{0.8258} & \textbf{0.9079} & 0.8752 & \textbf{0.9233} & \textbf{0.3666} & 0.7791 \\
& AdvS-SmartStu & 0.8205 & 0.9063 & \textbf{0.9038} & 0.9203 & 0.3622 & \textbf{0.8030} \\
\hline
\end{tabular}

\end{table}

\begin{table}[t]
\caption{Performance comparison when distilling from an ensemble of three teacher foundation models (UNI-v2, Phikon-v2, and Virchow-v2).}
\label{tab:table2}
\fontsize{8pt}{8.6pt}\selectfont
\centering

\setlength{\tabcolsep}{0.65mm}

\begin{tabular}{@{}l l cc cc cc@{}}
\hline
\multicolumn{2}{c}{} & \multicolumn{2}{c}{Yale HER2} & \multicolumn{2}{c}{SLN-Breast} & \multicolumn{2}{c}{BRACS} \\
\cline{3-8}
Model & Mode & bAcc & AUC & bAcc & AUC & bAcc & AUC \\
\hline
\multirow{3}{*}{\makecell{Original PFMs\\(300M $\sim$ 680M)}} &
UNI-v2 &
0.9032 & 0.9500 &
\cellcolor{red!12}\textbf{0.9323} & 0.9805 &
\textbf{0.5083} & 0.8699 \\
& Phikon-v2 &
\cellcolor{red!12}\textbf{0.9082} & \cellcolor{red!12}\textbf{0.9505} &
0.9271 & 0.9609 &
0.4527 & 0.8482 \\
& Virchow-v2 &
0.8629 & 0.9416 &
0.9203 & \textbf{0.9865} &
0.5059 & \cellcolor{red!12}\textbf{0.8747} \\
\hline
\multirow{4}{*}{\makecell{ResNet18\\(11.18M)}} &
Default &
0.7642 & 0.8479 &
0.7707 & 0.8722 &
0.3281 & 0.7455 \\
& Itg-SmartStu &
0.8366 & 0.9089 &
0.8895 & 0.9248 &
0.4298 & 0.8483 \\
& Adv-SmartStu &
0.8582 & 0.9153 &
0.9075 & 0.9383 &
0.4488 & \textbf{0.8573} \\
& AdvS-SmartStu &
\textbf{0.8674} & \textbf{0.9189} &
\textbf{0.9128} & \textbf{0.9549} &
\textbf{0.4576} & 0.8489 \\
\hline
\multirow{4}{*}{\makecell{ResNet34\\(21.28M)}} &
Default &
0.7787 & 0.8405 &
0.8030 & 0.8977 &
0.3471 & 0.7638 \\
& Itg-SmartStu &
0.8447 & 0.9337 &
0.8865 & 0.9519 &
0.4524 & 0.8553 \\
& Adv-SmartStu &
0.8605 & \textbf{0.9384} &
\textbf{0.9113} & \textbf{0.9684} &
0.4911 & 0.8574 \\
& AdvS-SmartStu &
\textbf{0.8708} & 0.9363 &
0.8955 & 0.9474 &
\textbf{0.5085} & \textbf{0.8596} \\
\hline
\multirow{4}{*}{\makecell{ResNet50\\(23.51M)}} &
Default &
0.8000 & 0.8758 &
0.8504 & 0.9098 &
0.3396 & 0.7150 \\
& Itg-SmartStu &
0.8616 & 0.9379 &
0.9090 & 0.9594 &
0.4699 & 0.8490 \\
& Adv-SmartStu &
0.8929 & 0.9384 &
0.9180 & 0.9609 &
0.4939 & 0.8482 \\
& AdvS-SmartStu &
\textbf{0.8976} & \textbf{0.9411} &
\textbf{0.9233} & \textbf{0.9820} &
\cellcolor{red!12}\textbf{0.5109} & \textbf{0.8591} \\
\hline
\multirow{4}{*}{\makecell{TinyViT-5M\\(5.07M)}} &
Default &
0.7513 & 0.8553 &
0.8323 & 0.8917 &
0.3497 & 0.7300 \\
& Itg-SmartStu &
0.8561 & 0.9326 &
0.9090 & 0.9293 &
0.4115 & 0.7922 \\
& Adv-SmartStu &
\textbf{0.8666} & 0.9321 &
0.9143 & 0.9459 &
0.4326 & 0.8076 \\
& AdvS-SmartStu &
0.8550 & \textbf{0.9342} &
\textbf{0.9180} & \textbf{0.9504} &
\textbf{0.4434} & \textbf{0.8118} \\
\hline
\multirow{4}{*}{\makecell{TinyViT-11M\\(10.55M)}} &
Default &
0.7871 & 0.8511 &
0.8271 & 0.8857 &
0.3664 & 0.7833 \\
& Itg-SmartStu &
0.8816 & 0.9332 &
0.9143 & 0.9564 &
0.4153 & 0.8328 \\
& Adv-SmartStu &
\textbf{0.8916} & 0.9395 &
0.9218 & 0.9429 &
0.4405 & 0.8361 \\
& AdvS-SmartStu &
\textbf{0.8916} & \textbf{0.9432} &
\textbf{0.9271} & \textbf{0.9699} &
\textbf{0.4902} & \textbf{0.8477} \\
\hline
\multirow{4}{*}{\makecell{TinyViT-21M\\(20.62M)}} &
Default &
0.7779 & 0.8268 &
0.8752 & 0.9398 &
0.3836 & 0.8028 \\
& Itg-SmartStu &
0.8924 & 0.9468 &
0.9180 & 0.9684 &
0.4367 & 0.8272 \\
& Adv-SmartStu &
\textbf{0.8974} & 0.9489 &
0.9233 & 0.9684 &
0.4607 & 0.8359 \\
& AdvS-SmartStu &
\textbf{0.8974} &
\cellcolor{red!12}\textbf{0.9505} &
\cellcolor{red!12}\textbf{0.9323} &
\cellcolor{red!12}\textbf{0.9940} &
\textbf{0.4958} & \textbf{0.8410} \\
\hline
\multirow{4}{*}{\makecell{MobileNetV3-Small\\(0.93M)}} &
Default &
0.7582 & 0.8474 &
0.8263 & 0.8827 &
0.3313 & 0.7414 \\
& Itg-SmartStu &
0.8147 & 0.8989 &
0.8805 & 0.9293 &
0.3657 & 0.7923 \\
& Adv-SmartStu &
\textbf{0.8303} & \textbf{0.9079} &
0.8805 & 0.9489 &
\textbf{0.3905} & 0.7914 \\
& AdvS-SmartStu &
0.8253 & 0.9061 &
\textbf{0.9090} & \textbf{0.9639} &
0.3886 & \textbf{0.8065} \\
\hline
\multirow{4}{*}{\makecell{MobileNetV3-Large\\(2.97M)}} &
Default &
0.7908 & 0.8579 &
0.8436 & 0.8782 &
0.3443 & 0.7511 \\
& Itg-SmartStu &
0.8205 & 0.9142 &
0.8985 & 0.9459 &
0.3978 & 0.8282 \\
& Adv-SmartStu &
\textbf{0.8671} & \textbf{0.9189} &
\textbf{0.9271} & \textbf{0.9474} &
\textbf{0.4162} & \textbf{0.8312} \\
& AdvS-SmartStu &
0.8363 & 0.9137 &
0.9090 & 0.9459 &
0.4062 & 0.8258 \\
\hline
\end{tabular}
\end{table}

\begin{figure}[t]
    \centering
    \includegraphics[width=\textwidth]{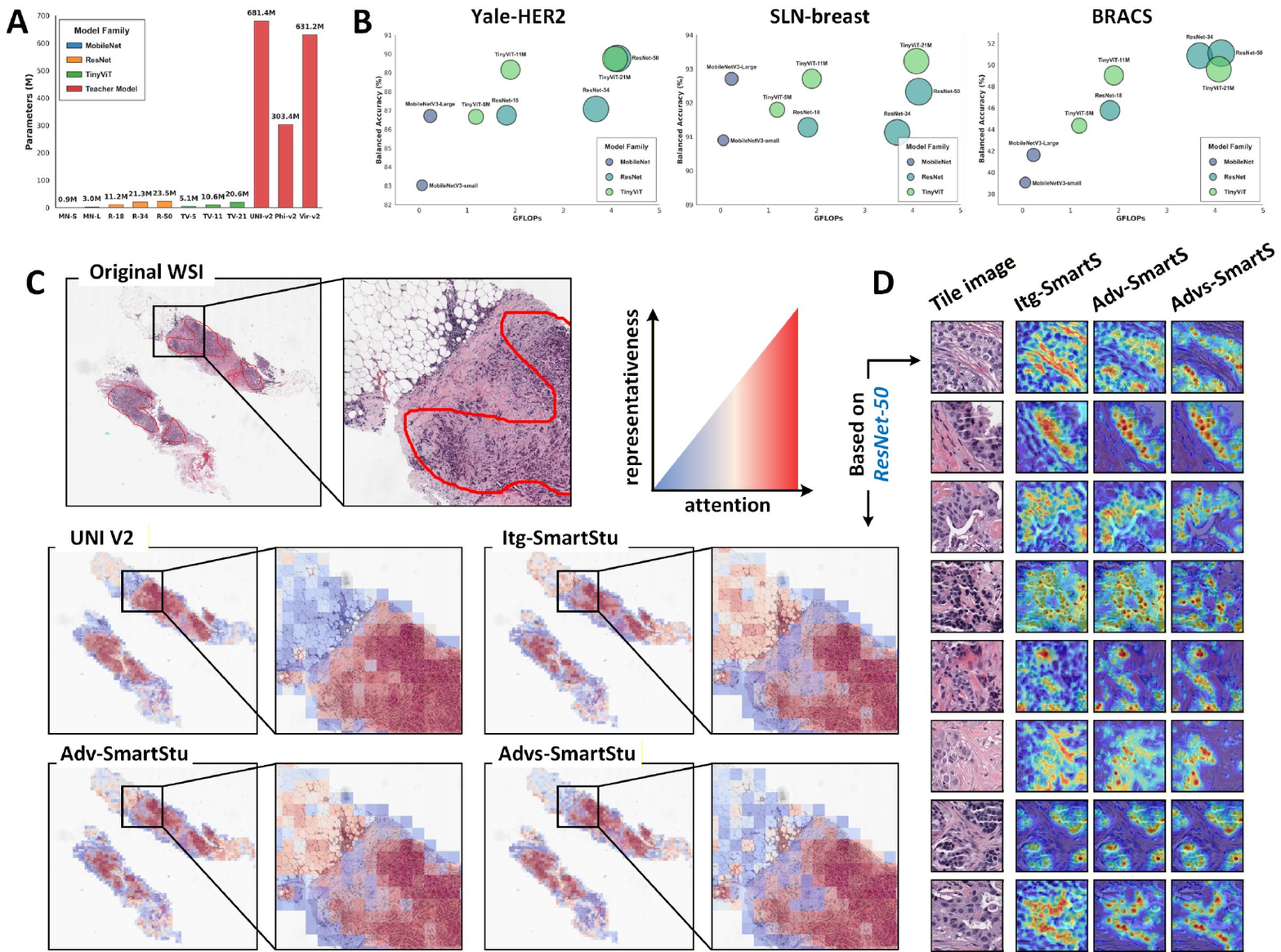}
    \caption{Efficiency and interpretability analysis of the proposed method. 
    \textbf{(A)} Parameter count comparison between student and teacher models. Abbreviations: MN (MobileNet-V3), R (ResNet), and TV (TinyViT). 
    \textbf{(B)} Performance-efficiency trade-off (Balanced Accuracy vs. GFLOPs), where bubble size is proportional to the parameter count. 
    \textbf{(C)} Slide-level attention heatmaps comparison on a sample from the Yale HER2 test set. 
    \textbf{(D)} Fine-grained interpretability visualization using tile-level Eigen-CAM.}
    \label{fig:result}
\end{figure}

\subsection{Datasets and implementation details}
\noindent\textbf{Datasets.} We pre-train SmartStu via distillation on TCGA-BRCA ($n=1{,}133$; 40 sites)\footnote{https://portal.gdc.cancer.gov/projects/TCGA-BRCA}. Generalisation is evaluated on three external breast cancer cohorts: Yale-HER2 ($n=192$) \cite{farahmand2022deep_YaleHER2,farahmand2022her2dataset_YaleHER2}, SLN-Breast ($n=130$) \cite{campanella2019slndataset_SLNbreast,campanella2019clinical_SLNbreast}, and BRACS ($n=547$) \cite{brancati2022bracs}. Yale-HER2 and SLN-Breast are binary classification tasks with a 0.8/0.2 train/test split. BRACS is a 7-class classification task, and we follow the dataset-default train/test split (0.84/0.16).

\noindent\textbf{Implementation.} WSIs are tiled into $256\times256$ patches from tissue regions. Preprocessing is performed with Trident toolbox~\cite{zhang2025standardizing_trident1}. Training has two stages: 1) adversarial teacher training with learning rate $1\times10^{-4}$, weight decay $0.05$, $\tau=0.07$, and $\lambda=1$; 2) student distillation with learning rate $1\times10^{-4}$, weight decay $0.05$, and loss weights $\lambda_{adv}=10^{-4}$, $\lambda_{ssl}=10^{-4}$.

For downstream evaluation, we employ ABMIL \cite{ilse2018attention_ABMIL} as the MIL aggregator. Our student models are compared against several PFMs, including UNI-V2 \cite{chen2024uni_uni}, Phikon-V2 \cite{filiot2024phikonv2largepublicfeature_phikonv2}, and Virchow-V2 \cite{zimmermann2024virchow2_virchowv2}. We evaluate a series of lightweight backbones for the student model, specifically ResNet (18, 34, 50), TinyViT (5M, 11M, 21M), and MobileNet-v3 (Small, Large). All experiments are conducted with 5 runs, with the average performance reported.
We employ balanced accuracy (BACC) and the area under the receiver operating characteristic curve (AUC) as the evaluation metrics.

In this context, \textbf{Default} denotes the model initialised with ImageNet\footnote{https://www.image-net.org} pre-trained weights. \textbf{Sig-SmartStu} and \textbf{Itg-SmartStu} denote single- and multi-teacher variants trained with distillation only (i.e., $\mathcal{L}_{distill}$). \textbf{Adv-SmartStu} is trained with $\mathcal{L}_{distill}+\lambda_{adv}\mathcal{L}_{adv}$. Finally, \textbf{AdvS-SmartStu} is trained with the full objective $\mathcal{L}_{all}$.

\subsection{Distillation performance results}
\noindent\textbf{Single-teacher setting.}
We first distil from UNI-v2 as the sole teacher (Table~\ref{tab:table1}). Compared with \textit{Default} (ImageNet pre-trained), distillation improves downstream performance, confirming successful knowledge transfer to compact backbones. Although these students generally remain below using frozen UNI features directly, adding adversarial distillation and then AdvS-SmartStu with the full objective still produces a clear, stepwise improvement, indicating that each component contributes to stronger cross-cohort generalisation.

\noindent\textbf{Multi-teacher setting.}
We next distil from an ensemble of three teachers (UNI-v2, Phikon-v2 and Virchow-v2). Table~\ref{tab:table2} shows a consistent progression: Itg-SmartStu (multi-teacher distillation only) improves over Default, and Adv-SmartStu further improves by explicitly discouraging site-related confounders. AdvS-SmartStu is commonly best or competitive across student families, suggesting that noise-guided de-biasing and artefact-augmented self-supervision encourage features that are less sensitive to acquisition-specific signatures while retaining diagnostically relevant morphology. However, very small backbones do not always benefit from the auxiliary self-supervised term, which is consistent with limited capacity and stronger regularisation pressure.

Across external cohorts (Table~\ref{tab:table2}), compact students can match or slightly exceed teachers in various backbone settings. This indicates that SmartStu can consolidate multiple teachers’ knowledge into a lightweight encoder while suppressing site-related bias via adversarial distillation; the auxiliary artefact-injection self-supervision further stabilises the learned features. Overall, these results support the portability of SmartStu under different deployment constraints.

\subsection{Efficiency and interpretable visualisation}
Figure~\ref{fig:result}-A shows that SmartStu students require substantially fewer parameters than the teacher PFMs, improving deployability in resource-constrained circumstances. Figure~\ref{fig:result}-B illustrates the relationship between performance and compute (BACC vs.\ GFLOPs). Overall, SmartStu concentrates strong performance in the low-compute regime, and several student architectures (notably ResNet-50 and TinyViT-21M) remain competitive across cohorts whilst operating at tens-fold lower computational cost than the teachers.

Figure~\ref{fig:result}-C compares tissue-level attention on a test WSI, where red contours mark tumour region of interest (ROI) annotations. In this example, the teacher PFM exhibits spurious activation outside the annotated ROI. SmartStu suppresses these false responses and concentrates attention on the ROI, improving localisation. AdvS-SmartStu further reduces background activation, producing a cleaner and more specific attention map on tumour regions.

Figure~\ref{fig:result}-D shows tile-level interpretation visualised by Eigen-CAM~\cite{muhammad2020eigen}. Activations become progressively more localised from Itg-SmartStu to AdvS-SmartStu, aligning better with salient morphology (e.g., nuclear boundaries and histological textures) whilst suppressing responses in less informative regions.

\section{Conclusion}
We propose \textbf{SmartStu}, a framework for customising compact, breast-cancer pathology foundation models. SmartStu introduces adversarial distillation to suppress site-related bias whilst retaining disease-relevant morphology, and incorporates multi-teacher distillation with artefact-injected self-supervision to stabilise tile representations. Across external cohorts, SmartStu substantially reduces model size and compute whilst maintaining strong downstream performance.
More broadly, adversarial distillation offers a controllable customisation paradigm for PFMs: by varying or combining nuisance targets, one can discourage specific non-biological signatures during knowledge transfer to better match personalised deployment. Future work will explore richer nuisance targets and adaptive combinations of adversarial objectives across clinical settings.

\begin{credits}
\subsubsection{\ackname} This work was supported by the following grants. JR is supported by the NIHR Oxford Biomedical Research Centre. HC is supported by the HKUST 30 for 30 Research Initiative Scheme (Project No. FS111). KY is supported by the National Natural Science Foundation of China (No. 62476101) and the Guangdong Basic and Applied Basic Research Foundation (Grant No. 2024A1515140137). 
The views expressed are those of the authors and not necessarily those of the NHS, the NIHR, or the Department of Health.

\end{credits}


\bibliographystyle{splncs04}
\bibliography{refs-0214}

\end{document}